\documentclass[journal,10pt]{IEEEtran}

\usepackage{ragged2e}
\usepackage{cite}
\usepackage{booktabs}
\usepackage{float}
\usepackage{graphicx}
\usepackage{subfigure}
\usepackage{color}
\usepackage{epstopdf}
\usepackage{amsthm,amsmath,amssymb,amsfonts,lipsum}
\usepackage{cases}
\usepackage{mathrsfs}
\usepackage{setspace}
\usepackage[ruled,linesnumbered,vlined]{algorithm2e}
\usepackage{stfloats}
\usepackage{enumerate}
\usepackage{amssymb}
\usepackage{bbm, dsfont}
\usepackage{multirow}
\usepackage[table]{xcolor}
\usepackage{bbm}
\usepackage{bm}
\usepackage{array}
\usepackage{tabularx}
\newcolumntype{C}[1]{>{\centering\arraybackslash}m{#1}} 
\newcolumntype{Y}{>{\centering\arraybackslash}X}        
\definecolor{tblBlue}{HTML}{D9E3F5}
\definecolor{tblGray}{HTML}{F2F2F2}
\usepackage{hyperref}
\allowdisplaybreaks[4]

\makeatletter
\newcommand{\rmnum}[1]{\romannumeral #1}
\newcommand{\Rmnum}[1]{\expandafter\@slowromancap\romannumeral #1@}

\makeatother

\begin{document}

\title{Large Multimodal Models for Embodied Intelligent Driving: Enabling the Next Generation of Autonomous Mobility}

\author{Long~Zhang,~\IEEEmembership{Member,~IEEE,}
	    Yuchen~Xia,
	    Bingqing~Wei,
	    Zhen~Liu,
	    Shiwen~Mao,~\IEEEmembership{Fellow,~IEEE,}
        Zhu~Han,~\IEEEmembership{Fellow,~IEEE}
        and~Mohsen~Guizani,~\IEEEmembership{Fellow,~IEEE}
\thanks{Long Zhang, Yuchen Xia, and Zhen Liu are with the School of Information and Electrical Engineering, Hebei University of Engineering, Handan 056038, China (e-mail: lzhang0310@gmail.com; xyc2000925@gmail.com; liuzhen@hebeu.edu.cn).}
\thanks{Bingqing Wei is with the School of Information Science and Engineering, Lanzhou University, Lanzhou 730000, China (e-mail: bingqingwei@hotmail.com).}
\thanks{Shiwen Mao is with the Department of Electrical and Computer Engineering, Auburn University, Auburn, AL 36849, USA (e-mail: smao@ieee.org).}
\thanks{Zhu Han is with the Department of Electrical and Computer Engineering, University of Houston, Houston, TX 77004, USA (e-mail: hanzhu22@gmail.com).}
\thanks{Mohsen Guizani is with the Machine Learning Department, Mohamed Bin Zayed University of Artificial Intelligence, Abu Dhabi 99163, UAE (email: mguizani@ieee.org).}
}

\maketitle

\begin{abstract}
The advent of Large Multimodal Models (LMMs) offers a promising technology to tackle the limitations of modular design in autonomous driving, which often falters in open-world scenarios requiring sustained environmental understanding and logical reasoning. Besides, embodied artificial intelligence facilitates policy optimization through closed-loop interactions to achieve the continuous learning capability, thereby advancing autonomous driving toward embodied intelligent (EI) driving. However, such capability will be constrained by relying solely on LMMs to enhance EI driving without joint decision-making. This article proposes a novel semantics and policy dual-driven hybrid decision framework to tackle this challenge, ensuring continuous learning and joint decision. The framework merges LMMs for semantic understanding and cognitive representation, and deep reinforcement learning for real-time policy optimization. This article particularly details the foundational principles of EI driving and LMMs, and also introduces the emerging opportunities this framework enables, encompassing potential benefits and representative use cases. The proposed framework, evaluated through a lane-change planning case study, achieves at least a 19\% performance improvement over the baselines, as verified by simulations. Finally, key open challenges and future research directions are identified to guide subsequent work.
\end{abstract}


\IEEEpeerreviewmaketitle

\vspace{-0.3cm}
\section{Introduction}
Autonomous driving, which represents the new generation of intelligent transportation capable of significantly improving safety and operational efficiency, has advanced rapidly in both academia and industry\cite{ZhangL2025}. Academic research has evolved from conventional modular architectures toward end-to-end learning and foundation model-enabled driving. Meanwhile, industrial systems have progressed from widely deployed driver-assistance functions to higher-level autonomous driving under restricted operating conditions. Despite these advances, existing autonomous driving systems remain limited by insufficient sustained environmental understanding and logical reasoning capabilities\cite{ClaussmannL2019}, which constrain their reliability, safety, and generalization in complex real-world traffic scenarios. Consequently, further investigation is imperative to refine their understanding, reasoning, and adaptive capabilities.

The field of artificial intelligence (AI) has been revolutionized by Large Language Models (LLMs) and the recently emerged Large Multimodal Models (LMMs) \cite{ChenL2024}. In response to the above limitations, LMMs have become an essential technology to improve the environmental understanding and decision-making ability of autonomous driving systems. By interpreting multimodal driving data via LMMs, autonomous vehicles (AVs) acquire a thorough semantic understanding of complex scenes, and perform behavioral prediction \cite{HuangS2025}, thereby enhancing their safety and reliability in real-world operations. Several recent works on LMM-empowered autonomous driving have been explored from the different perspectives, such as external environment understanding \cite{LiaoH2025}, driving decision and control \cite{XuZ2024}, driver state recognition \cite{HuC2025}, etc. However, as the traffic environment becomes increasingly complex, these capabilities presented by LMMs struggle to support autonomous adaptation and evolution in real-time interaction between AVs and their environment under dynamic conditions. Therefore, more powerful methods are needed to address the growing complexity of the traffic environment.

In this context, embodied AI has emerged as a revolutionary means of promoting autonomy in autonomous driving systems. Unlike conventional AI models, embodied AI emphasizes the interactions between embodied agents and their physical environment \cite{DuanJ2022}. By embedding embodied AI, AVs evolve into embodied intelligent vehicles (EIVs), enabling them to adapt to dynamic environmental changes and make intelligent decisions from interactive data. Such adaptability, in turn, allows them to continuously refine their driving policies through environmental interaction, thereby materializing the embodied intelligent (EI) driving paradigm. Currently, embodied AI is in the exploratory stage, used for real-time decision and action planning in dynamic and uncertain environments. Great potential has been particularly demonstrated in the transportation-related scenarios, including vehicular networks and rail transportation systems \cite{ZhangR2025,ChenM2025,ZhouM2024,LiL2025}.

\begin{table*}[tb]
	\footnotesize
	\centering
	\caption{Comparison with the existing works on LMMs and DRL in the autonomous driving systems and other transportation-related scenarios.}
	\label{table-Simulation Parameters}
	\begin{tabularx}{\textwidth}{|
			>{\columncolor{tblBlue}\centering\arraybackslash}m{1.8cm}|
			>{\columncolor{tblGray}\centering\arraybackslash}m{0.7cm}|
			>{\columncolor{tblBlue}\centering\arraybackslash}m{0.8cm}|
			>{\columncolor{tblGray}\centering\arraybackslash}m{0.7cm}|
			>{\columncolor{tblBlue}\centering\arraybackslash}m{1.1cm}|
			>{\columncolor{tblGray}\raggedright\arraybackslash}X|}
		\hline
		\textbf{Reference} & \textbf{Year} & {\textbf{LMMs}} & \textbf{DRL} &\textbf{Joint Decision} & \multicolumn{1}{>{\columncolor{tblGray}\centering\arraybackslash}c|}{\textbf{Main Contributions}}\\
		\hline
		Liao et al.\cite{LiaoH2025} & $2025$ & $\checkmark$ & $\times$ & $\times$  & They proposed an LMM-based traffic accident anticipation framework through the CoT prompting. By hierarchically fusing video, optical flow, and linguistic descriptions, the early perception capability can be effectively enhanced for high-risk scenarios. \\
		\hline
		Xu et al.\cite{XuZ2024} & $2024$ & $\checkmark$ & $\times$ & $\times$ & They developed an interpretable end-to-end autonomous driving system based on the visually instruction-tuned and mixed fine-tuned LMMs. This system achieves the unified modeling from forward-view videos to low-level control commands and natural language explanations.\\
		\hline
		Hu et al.\cite{HuC2025} & $2025$ & $\checkmark$ & $\times$  & $\times$  & They introduced a training-free multimodal LLM method driven by human-centric context and self-uncertainty. This method extracts the scene graphs and integrates the multiple reasoning responses through the evidence theory to recognize the driver states accurately without any training.\\
		\hline
		Zhang et al.\cite{ZhangR2025} & $2025$ & $\checkmark$ & $\checkmark$  & $\times$ & They designed an embodied AI-enhanced vehicular network framework that integrates the vision-LLMs for semantic extraction and the DRL to stabilize decision under uncertainty. This framework significantly enhances the efficiency of semantic communication via the vision-LLMs and decision through the DRL.\\
		\hline
		Chen et al.\cite{ChenM2025} & $2025$ & $\checkmark$ & $\times$  & $\times$ & They presented an embodied AI‑enabled perception and decision framework for the scenarios of connected vehicles. By employing LMMs, this framework allows vehicle-to-vehicle (V2V) data to achieve greater situational awareness and reliable decision in uncertain scenarios.\\
		\hline
		Zhou et al.\cite{ZhouM2024} & $2024$ & $\checkmark$ & $\checkmark$  & $\times$ & They proposed a framework for an autonomous operation of rail transportation system enabled by embodied AI through integrated perception, execution, and learning. This framework allows the system to improve its adaptability to the environment and enhance autonomous decision capabilities.\\
		\hline
		Li et al.\cite{LiL2025} & $2024$ & $\checkmark$ & $\times$  & $\times$ & They introduced an LMM-empowered embodied AI framework for autonomous driving in mining. By incorporating the interactive learning and CoT reasoning, the framework can improve the decision interpretability and environmental adaptability of mining vehicles. \\
		\hline
		This article & --- & $\checkmark$ & $\checkmark$ & $\checkmark$ & We propose a joint decision framework by merging the LMMs and DRL for EI driving. Our framework particularly combines the multimodal semantic reasoning with closed-loop policy optimization. Such combination enhances adaptation to the dynamic driving environment and the overall generalization. \\
		\hline
	\end{tabularx}
	\vspace{-0.1cm}
\end{table*}

Although these achievements are of great significance, as shown in Table \ref{table-Simulation Parameters}, most of them emphasize either multimodal semantic reasoning or deep reinforcement learning (DRL)-based policy optimization. The coupling between semantic reasoning and executable policy learning remains insufficiently explored, and a closed-loop design mechanism capable of continuous learning in response to environmental changes within dynamic traffic conditions has yet to be established. While LMMs enhance understanding of multimodal driving data, DRL is required to facilitate joint decision in dynamic environments. The decision process often necessitates continuous interactions between the agents and their environment. Therefore, incorporating the continuous learning and joint decision in dynamic scenarios into EI driving remains an emergent research frontier.

\begin{figure*}[t]
	\centering
	\includegraphics[width=6.9in]{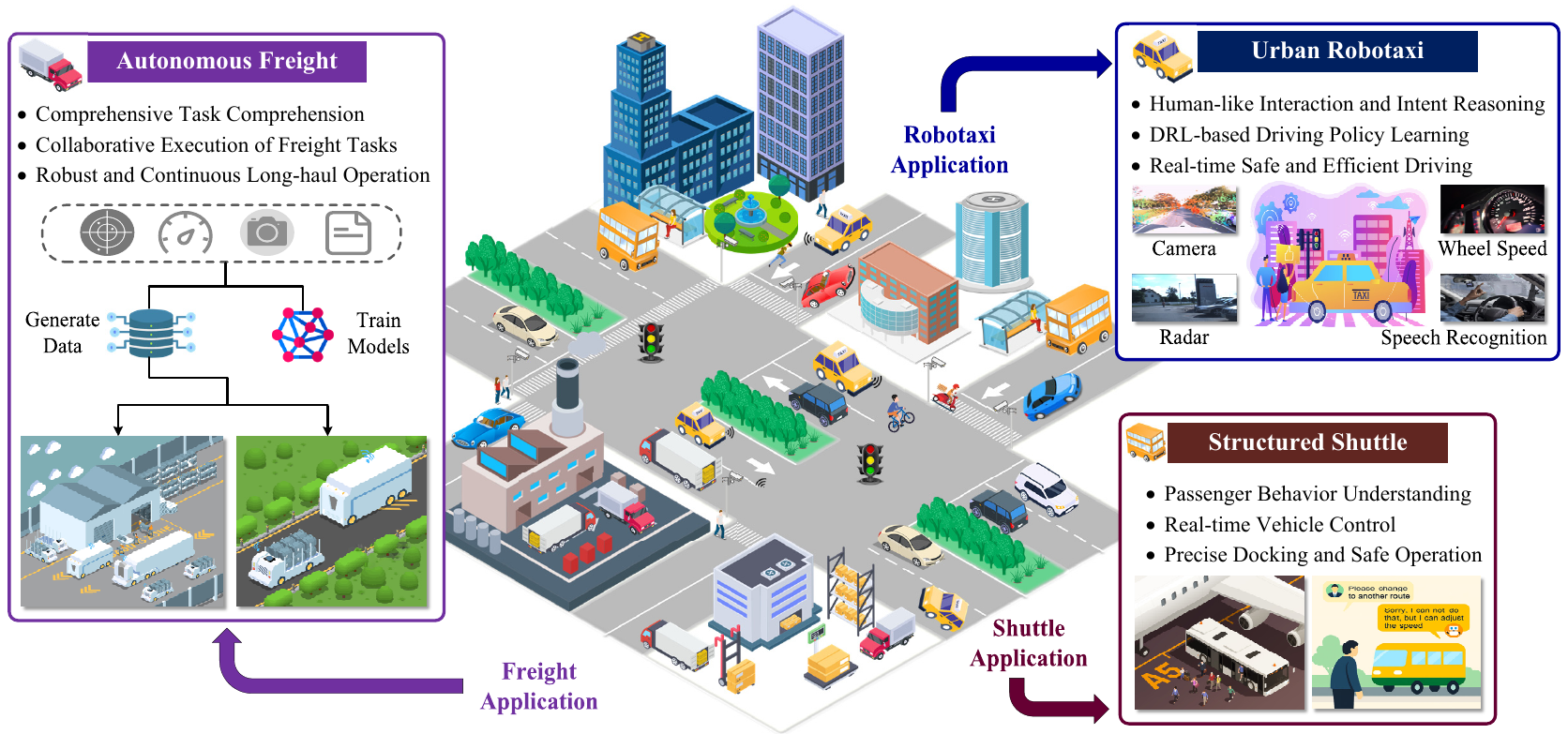}
	\caption{Illustration of an EI driving environment enabled by the joint decision approach based on LMMs and DRL. Such approach shows great promise in three illustrative application scenarios: \rmnum{1}) Urban robotaxi, which is on-demand and driverless transport, faces tremendous challenges in negotiating highly dynamic traffic conditions. Examples would be predicting a pedestrian's intent, and avoiding a sudden obstacle in a narrow corridor. \rmnum{2}) Autonomous freight operates in trunk-line logistics, ports, and industrial parks, where good performance is needed in complex geometrical road situations. Representative scenarios include container handling, and narrow space delivery robots. \rmnum{3}) Structured shuttle, driving in structured and semi-open environments, must achieve precision docking. Typical scenarios include airport shuttle service, and micro-circulation bus.}
	\label{Fig-intro-applications}
\end{figure*}

In this article, we propose employing the joint decision based on LMMs and DRL to enhance EI driving. LMMs inherit and extend the capabilities of LLMs, traditionally designed for textual data processing. Through architectural adaptation and multimodal collaborative training via LMMs, EIVs effectively handle the collected multimodal data \cite{ChengL2025}. These data processed and integrated into internal representation thereby enhance its generalization ability to new environments and tasks. In contrast, DRL behaviors can be adjusted by real-time feedback, which results in adaptive policy development in dynamic operations \cite{ZhangR2023,ZhangL2025B}. In DRL, the EIV interacts with the environment and accordingly, receives feedback in the form of rewards or penalties, by which it learns to make optimal decision. We will now demonstrate the application of LMMs and DRL-based joint decision across various scenarios, as shown in Fig. \ref{Fig-intro-applications}. The contributions of this article are summarized below.

\textbf{Fundamental Concepts and Requirements.}
We introduce the foundational concept of EI driving, highlighting its closed-loop pipeline of perception, decision, and execution as a core architectural design for EIVs. A brief outline of the basic architecture of LMMs, including modules and implementation process, is provided. Given such fundamentals, we demonstrate the requirements for integrating LMMs with EI driving.

\textbf{Framework and Opportunities.}
To implement the LMM-enhanced EI driving, we propose a semantics and policy dual-driven hybrid decision framework, which integrates LMMs, for semantic understanding and cognitive representation, and DRL, for real-time policy optimization. Additionally, we identify the emerging opportunities this framework enables, including the potential benefits and transformative use cases.

\textbf{Case Study and Evaluations.}
We evaluate the proposed framework in a case study of lane-change planning through simulations, utilizing the scenario mixed with an EIV and multiple conventional human-driven vehicles (HVs). The simulation results indicate that our framework achieves superior performance in both policy quality and dynamic adaptability.

\section{Overview of LMMs and EI Driving}
This section begins with the fundamental concept of EI driving, followed by the general framework of LMMs. Then, the necessities of integrating EI driving and LMMs are discussed.

\subsection{Fundamentals of EI Driving}
Embodied AI refers to the intelligent behavior generated by the interaction between embodied agent and its external environment \cite{DuanJ2022}. Such behavior emphasizes the direct physical world interaction, and the close coupling among perception, action, and learning. By integrating embodied AI, embodied agent obtains human-like environmental perception and real-time feedback. This enables the embodied agent to demonstrate the capabilities of continuous learning and autonomous evolution through its morphological characteristics and dynamic interaction with the environment.

\begin{figure}[t]
	\centering
	\includegraphics[width=3.6in]{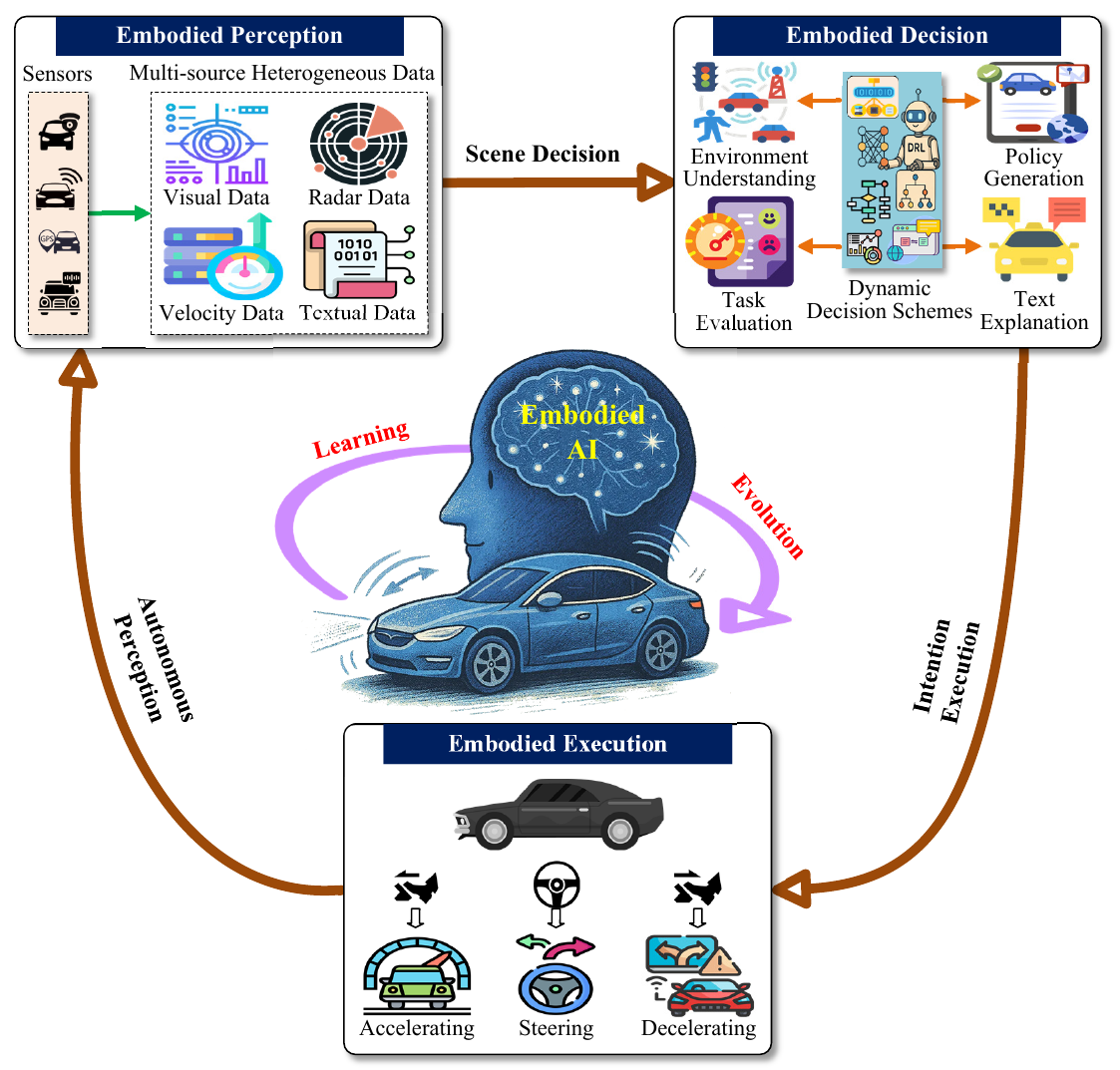}
	\caption{The architecture of EIV, which consists of the embodied perception module, embodied decision module, and embodied execution module. The EIV first achieves the autonomous environmental perception through multimodal sensors, and collects heterogeneous data sources, such as vision, radar, velocity, and textual data. Based on the comprehensively fused perception, the EIV proceeds to interpret the surrounding environment and perform the scenario-based decision, thereby formulating the high-level policy intentions. Finally, these policy intentions are translated into the concrete control actions for EIV. These modules operate as a closed-loop pipeline, including the perception, decision, and execution.}
	\label{Fig-Architecture-EI vehicle}
\end{figure}

The integration of embodied AI into an EIV enhances its ability to comprehensively understand all interactive behaviors, thereby showing a higher degree of intelligence. By continuously collecting multimodal data through environment interaction and feeding the data back to the EIV, driving maneuver commands are rapidly generated and executed, enabling real-time adaptation to dynamic environmental conditions. As shown in Fig. \ref{Fig-Architecture-EI vehicle}, the architecture of EIV adopts a closed-loop structure of perception, decision, and execution \cite{ZhouM2024}, corresponding to the following three modules.

\textbf{Embodied Perception.}
Based on onboard sensors, this module collects high-fidelity, multimodal raw data from physical world, endowing EIV with the capability for active understanding of complex environments. The collected data include four modalities: vision, radar, velocity, and text. The visual modality utilizes onboard camera to acquire the environmental images, which facilitates the perception of traffic elements, e.g., roads, obstacles, and traffic signs. The radar modality uses the LiDAR and radar sensors to achieve precision detection of dynamic objects. The velocity modality integrates the GPS and wheel-speed sensors to obtain real-time position and motion states of EIV. The textual modality analyzes the system internal information and interprets the driver's instructions.
	
\textbf{Embodied Decision.}
As a modular core of EIV, this module efficiently converts perceptual results into the executable behavioral policies. By continuously evaluating real-time traffic conditions, it dynamically adjusts its policy to jointly optimize the cognition and decision. Through various intelligent decision methods, e.g., LMMs and DRL algorithms, the dynamic policy coupled with EIV's physical properties and the external environment is generated. Real-time evaluations of task requirements, environmental changes, and vehicular capabilities are performed. This enables the flexible responses to emergencies, and the formulation of safe yet operationally efficient action plans. The mechanism of self-learning and optimization is also integrated into the module, realizing the continuous evolution during actual operation. The adaptability and overall intelligence level are thereby enhanced for decision process.
	
\textbf{Embodied Execution.}
This module translates executable decisions into precise driving maneuver commands, which embodies precision control and action execution. Aided by the closed-loop coordination with the embodied perception and decision modules, maneuver command dispatching like driving, steering and braking can be accurate. Besides, the module highlights adjusting actions and the external environment with feedback in real-time. The actuation system of EIV should be highly responsive and operationally robust. The feedback data enables dynamic fine-tuning and optimization of the execution policies, thus guaranteeing smooth motion and safe operation.

\begin{figure*}[t]
	\centering
	\includegraphics[width=5.0in]{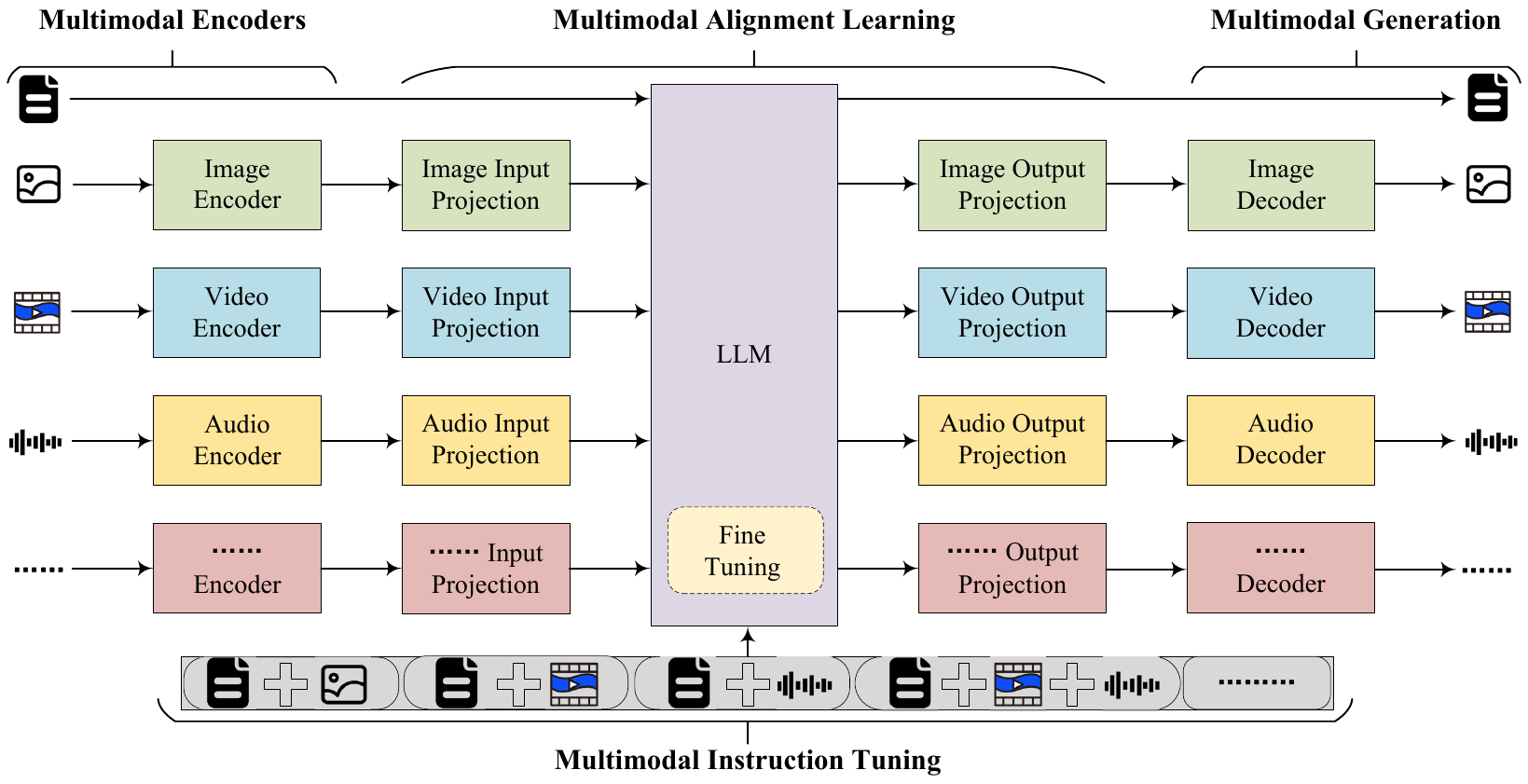}
	\caption{An overview of the overall LMM framework. The LMMs achieve unified cross-modal understanding and support any-to-any modality inputs and outputs through multimodal encoders, multimodal alignment learning, multimodal generation, and multimodal instruction tuning.}
	\label{Fig-LMM-flowchart}
\end{figure*}

\subsection{Fundamentals of LMMs}
LLMs learn from massive amounts of text to consider global information within input sequences and target the most relevant parts. LMMs utilize LLMs as their cognitive core, by integrating the modality-specific encoders and various decoders \cite{ChengL2025}. This thereby achieves the heterogeneous multi-source data processing and the generation of multimodal outputs. Notably, LMMs adopt a two-stage training method: multimodal pretraining and fine-tuning. Pretraining stage bridges the semantic gap between multimodal features and textual features to achieve cross-modal alignment. Fine-tuning stage further enhances their capabilities to understand and execute complex instructions. As shown in Fig. \ref{Fig-LMM-flowchart}, the LMM framework includes the following key components.

\textbf{Multimodal Encoders.}
Multimodal encoding aims to extract discriminative features from heterogeneous input data. For example, the encoders can extract the visual features from visual modalities (e.g., images or videos), and the acoustic features from auditory modalities (e.g., audio waveforms). These features are converted into a unified representation space to achieve effective interaction between cross-modal information. Due to the differences across various data modalities, the features of each modality are encoded differently in the representation space. Feature alignment learning is therefore required to address this problem.

\textbf{Multimodal Alignment Learning.}
Multimodal alignment learning is important to eliminate cross-modal semantic gap. The multi-head attention mechanism captures discriminative features in different modal subspaces. At each level of attention, certain modality-specific features are captured, which are then integrated into their own higher representation space, thereby making them cross-modally consistent.

\textbf{Multimodal Generation.}
Multimodal generation seeks to effectuate a dynamic composition between input and output modalities. Advanced generative methods are increasingly being used in LMMs, including Transformers and diffusion models. Different input data modalities yield semantic information. The cross modal outputs are then generated in any modality combination. 

\textbf{Multimodal Instruction Tuning.}
Multimodal instruction tuning aims to comprehend and carry out cross-modal instructions, which refers to building the pairs of instruction semantics to the output data. LMMs apply the In-Context Learning (ICL) and Chain-of-Thought (CoT) reasoning capabilities in a multimodal environment, enhancing their cross-modal generalization performance. In complex and varied cross-modal tasks, performance enhancement is especially observed.

\subsection{Necessities for Integrating EI Driving and LMMs}
EI driving operates in highly complex traffic environments, characterized by heterogeneous multimodal data, multi-scale spatial structures, and dynamic human-machine interactions. The cross-modal understanding and task generalization capabilities of LMMs offer a consolidated groundwork for perception and decision at EIV, thereby acting as a vital technological pillar for the realization of EI driving.

\textbf{Semantic Understanding of Complex Traffic Scenarios.}
One of the critical challenges in EI driving is how to process heterogeneous multimodal data. Enabled by the cross-modal embedding and semantic unification, LMMs address this challenge by projecting diverse data modalities into a coherent cognitive representation space. This significantly enhances the ability of EI driving system to understand and reason about semantic relationship among environmental elements.

\textbf{Human-Vehicle Collaboration and Teaming.}
LMMs can jointly model the natural language, visual behavioral signals, and traffic context. EI driving system thereby infers the intentions of drivers, pedestrians, and passengers through the established semantic relationship. This enables the EIV to interpret and explain its own behavior. With LMMs, the system can respond to the driving maneuver commands and predict the actions of interactive embodied agent, thereby facilitating operationally efficient and interactionally fluent human-machine cooperative driving.

\textbf{Human-Like Decision in Dynamic Interaction.}
EI driving system needs to make decision under highly uncertain and rapidly evolving environments. By integrating historical motion trajectories, environmental states, and contextual cues, LMMs perform the execution of CoT reasoning to generate higher-level policies, including both intent understanding and trajectory prediction. The flexible decision capability empowers the system to adapt to complex scenarios, thereby advancing its level of generalizable intelligence.

\begin{figure*}[t]
	\centering
	\includegraphics[width=5.6in]{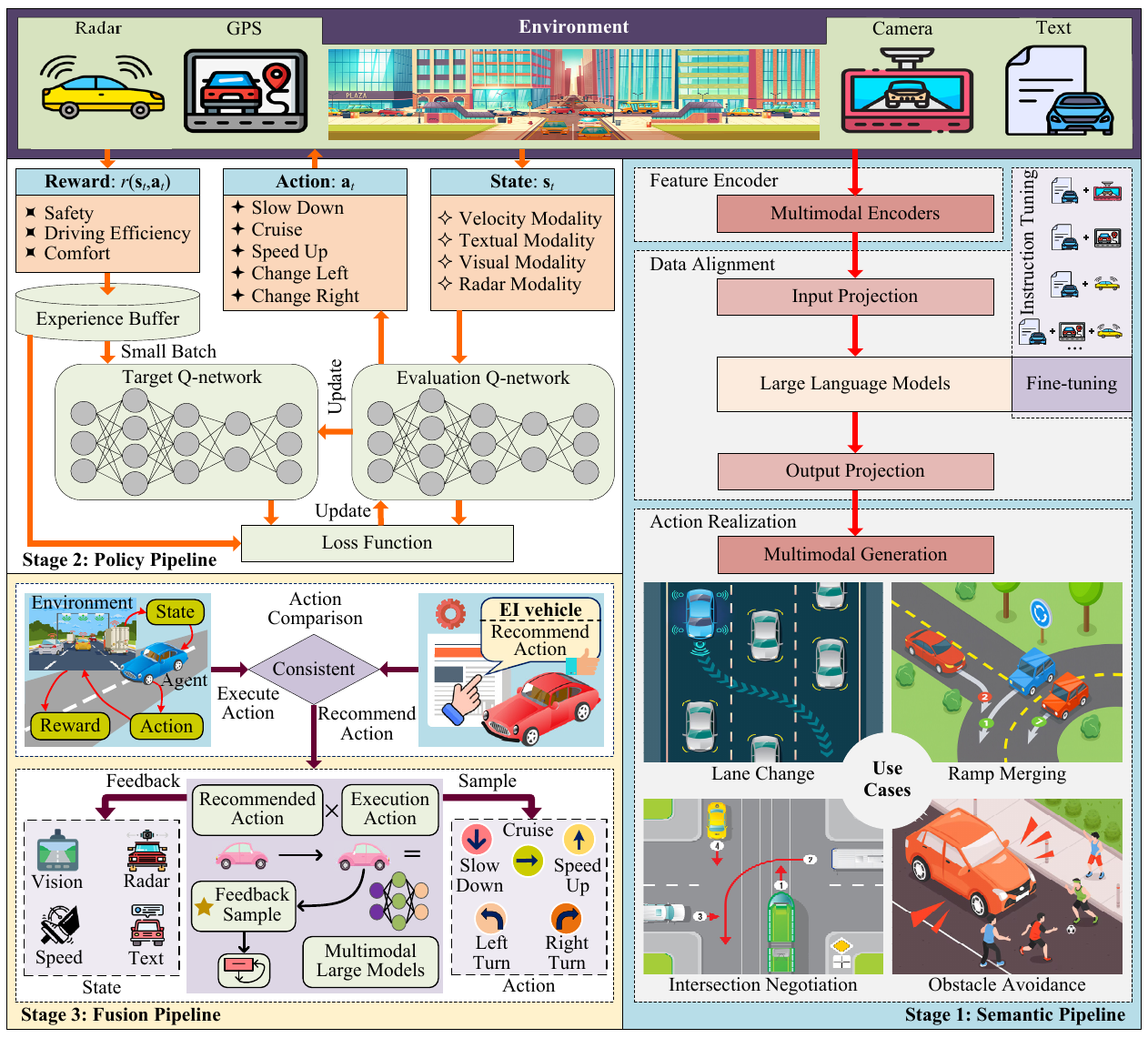}
	\caption{Illustration of the proposed semantics and policy dual-driven hybrid decision framework to implement the LMM-empowered EI driving. Three successive stages are designed and incorporated into this framework: semantic pipeline, policy pipeline, and fusion pipeline.}
	\label{Fig-framework}
\end{figure*}

\section{Integration of LMMs and EI Driving}
This section proposes a procedural framework for implementing the LMM-empowered EI driving. The emerging opportunities this framework enables are also introduced, including the potential benefits and promising use cases.

\subsection{Framework}
To implement the LMM-enabled EI driving, we propose a semantics and policy dual-driven hybrid decision framework, as shown in Fig. \ref{Fig-framework}. The proposed framework integrates DRL, known for its high responsiveness and policy adaptability in complex dynamic environments, with LMMs, which exhibit exceptional capability in semantic reasoning and high-level decision representation. The combination of these complementary strengths empowers the EI driving system to understand task context and adapt to dynamic environmental conditions.

\textbf{Stage 1. Semantic Pipeline.}
The semantic pipeline leverages LMMs to obtain the higher-level driving action policy, through the CoT processing of unified multimodal cognitive representation. The procedure of this pipeline is specified below.

\begin{itemize}
	\item \textit{Step 1. Feature Encoding:}
	This step performs the semantic-level encoding operations on raw multimodal data. The inputs from diverse sources, including vision, radar, semantic annotations, and communications, are transformed into the machine interpretable semantic feature representations. The step preserves modality-specific structural information and establishes preliminary cross-modal association. The semantic relationship between different modalities can be thereby learned.
	
	\item \textit{Step 2. Data Alignment:} 	
	Firstly, the cross-modal projection constructs a feature input projection network to achieve consistent mapping of non-linguistic modalities, e.g., images, velocity, and radar signals, into a unified semantic space. This step ensures that all heterogeneous data are coherently represented within a unified cognitive representation space. Then, LLMs comprehensively analyze the driving scenario by processing text commands and multimodal data. Text outputs and signal tokens are generated for control and decision, including path planning and behavior prediction. Finally, the output projection maps signal tokens to specific modal demand space, used for the multimodal content generation.
	
	\item \textit{Step 3. Action Realization:} 
	This step transforms various signal tokens into the decisions and predictions from multimodal data, to guide the action commands of EIV. By converting high-level semantic intents into specific physical actions, visual interface prompts and voice interaction feedback are used to ensure system interpretability and facilitate human-centered interactivity.
	
	\item \textit{Step 4. Instruction Tuning:} 
	This step acquires an understanding of the policy intentions through a limited number of examples. CoT reasoning and ICL mechanisms facilitate hierarchical task decomposition, which progressively derives behavioral logic to enhance the precision of instruction generation. Through continuous reasoning, the EI driving system improves environmental adaptability and task execution stability.
\end{itemize}

\textbf{Stage 2. Policy Pipeline.}
To ensure real-time performance optimization of EIV in dynamic and open environments, the policy pipeline employs DRL approach that directly maps the collected multimodal data into executable action policy. With DRL, the policy network parameters are continuously optimized through online interaction mechanism \cite{ZhangL2025}. These updates strategically integrate the recommended actions generated by semantic reasoning as the policy references for joint decision, thereby generating the execution action policy. This ensures the dynamic responsiveness to evolving environmental states, achieving stable and operationally efficient control behaviors. The DRL framework is designed as follows.

\begin{itemize}
	\item \textit{Markov Decision Process (MDP) Formulation:}
	The real-time action policy generation problem is formulated as an MDP, including the state space, action space, and reward. The state $\mathbf{s}$ consists of the collected multimodal data, including the EIV's own status, e.g., velocity and textual modalities, and the environmental information, e.g., visual and radar modalities. The action $\mathbf{a}$ is sampled from the advantage estimates over all feasible actions within the discrete action space, involving the driving maneuvers, e.g., turning left, turning right, proceeding straight, accelerating, decelerating, and maintaining constant speed. The reward is defined as the sum of all reward terms at time-step $t$, i.e., $R\left(\mathbf{s}_{t},\mathbf{a}_{t}\right) =R_{t}^{\textrm{sfty}} + R_{t}^{\textrm{de}} + R_{t}^{\textrm{comf}}$, including the safety, driving efficiency, and comfort terms. Safety term penalizes the lane-change behavior that jeopardizes driving safety. When the EIV collides with another vehicle or enters an unsafe terminal state, the current episode is terminated and a collision penalty of ${\delta_1}$ is imposed. Driving efficiency term, including speed and lane-change rewards. Vehicle's speed undergoes the linear normalization within a predefined target speed range. Specifically, the speed below the minimum threshold yields penalty, while the speed approaching the maximum value receives higher reward. When executing a lane-change maneuver under the environment safety constraint, EIV receives a reward of ${\delta_2}$. This reward encourages proactive and flexible driving policies, when slow moving vehicles are detected ahead. Comfort term provides a reward of ${\delta_3}$ when the EIV maintains a stable on-road state, thereby discouraging unnecessary lateral oscillations and frequent lane changes. These terms enable the EIV to learn an optimal policy $\pi^*$ that maximizes the expected return.
	
	\item \textit{Environment Interaction and Training Process:} 
	At each time-step $t$, EIV interacts with the environment and selects an action $\mathbf{a}_{t}$ from current state $\mathbf{s}_{t}$ based on a noisy network. Upon taking action $\mathbf{a}_{t}$, EIV receives a reward $R\left(\mathbf{s}_{t},\mathbf{a}_{t}\right)$ and the environment transits to next state $\mathbf{s}_{t+1}$. The transition is stored into the experience replay buffer, and current state $\mathbf{s}_{t}$ and next state $\mathbf{s}_{t+1}$ are fed into the evaluation network and target network, respectively, to obtain the Q-values for different actions. The action $\mathbf{a}_{t+1}$ corresponding to the maximum Q-value is selected from the evaluation network, and its Q-value is obtained from the target network. This value is then used to guide the training and update the evaluation network. The target network gradually adjusts its parameters through soft updates to reduce the instability during training.
\end{itemize}

\textbf{Stage 3. Fusion Pipeline.}
With the first two stages, the final execution action is obtained through the policy pipeline, while the semantics output from the semantic pipeline is referenced during the generation process. However, due to the differences in decision of semantic and policy pipelines, their generated results may not exhibit the complete consistency. Without effectively identifying such differences, LMMs may encounter difficulties in maintaining alignment with the actual policy during long-term reasoning. To address this issue, the fusion pipeline is introduced to enable the action consistency discrimination and facilitate the backward updating of the policy, enabling continuous learning.

\begin{itemize}
	\item \textit{Step 1. Action Comparison:} 
	This step compares the action executed by the policy pipeline with the reasoning output of the semantic pipeline under the same states. If the two outcomes are consistent, this step indicates that the semantic reasoning has successfully aligned with the policy preference, requiring no further adjustment. Otherwise, the discrepancy is identified and constructed as a feedback sample, referred to as a failure case. Each feedback sample contains the current multimodal information, the semanticlly recommended action, the final executed action, and the next state.
	
	\item \textit{Step 2. Feedback \& Adaptation:} 
	When action inconsistency is detected, the corresponding state-action pair is constructed as a feedback sample and returned to the semantic pipeline for periodic fine-tuning.  Specifically, the policy pipeline is updated frequently through environmental interaction, whereas the semantic pipeline is periodically updated using accumulated feedback samples. Low‑Rank Adaptation (LoRA)-based fine-tuning is adopted to refine the outputs of the semantic pipeline. In the short term, it helps ensure real-time response and operational stability during execution. Over the long term, this enforces deep coupling between the semantics and policy pipelines, which enhances the generalization ability of EI driving system in the complex environment.
\end{itemize}

\subsection{Benefits}
The proposed framework introduces the following potential benefits for the EI driving system.

\textbf{Enhanced Environmental Perception.}
The system uses multimodal heterogeneous data to predict road participants' behavioral intention and its ultimate environmental understanding. This cross-modal information processing capability of EIV transcends the static limitation of conventional perception, resulting in much better awareness of the environment.

\textbf{Improved Decision Reliability.}
The system dynamically adjusts the driving policy in real-time within dynamic environments, while rapidly identifying anomalous objects and executing strategic planning. This ensures robust state estimation and policy selection in complex, open-world scenarios, thereby enhancing the reliability of decision.

\textbf{Strengthened Continual Learning Capability.}
During the training process, the system converts action discrepancies into the learning signals, and fine-tunes them in combination with historical interactive data, enabling the policy adaptation to unknown scenarios during operation. The stronger environmental adaptability and improved sustainable learning capacity can be obtained.

\subsection{Use Cases}
Within the proposed framework, several promising use cases are illustrated below.

\textbf{Lane Change.}
As a challenging driving maneuver, lane-change requires EIV to maintain stable move within its original lane, while simultaneously selecting and entering a target lane. Our framework is capable of reasoning about the lane-changing feasibility and generating the specific lane-change planning actions. This thereby enables the safe and efficient lane-change execution within complex traffic flows.

\textbf{Ramp Merging.}
In the ramp merging scenarios, EIV must safely and smoothly merge into the main traffic flow within constrained spatial and temporal conditions. Our framework enables the accurate traffic scene interpretation and the merging opportunity identification. Real-time merging decisions can also be generated to achieve seamless integration with dynamic traffic flows.

\textbf{Intersection Negotiation.}
Intersection negotiation presents multiple interaction challenges for EI driving. Our framework can analyze traffic regulations, identify the intentions of other vehicles, e.g., HVs, and interpret pedestrian behaviors. Meanwhile, path planning and behavior coordination can be well performed within the road structure, thereby maintaining both traffic efficiency and safety in highly uncertain intersection environments.

\textbf{Obstacle Avoidance.}
Obstacle avoidance involves the rapid detection and response to static and dynamic obstacles. Our framework can precisely identify the obstacle locations and motion states, and obtain immediate avoidance actions. Both the continuity of vehicle motion and operational safety can be thereby ensured under extreme scenarios.

\section{A Case Study of Lane-Change Planning}

\textbf{Scenario and Baselines.}
We present a case study based on simulations to evaluate the performance of our framework, and consider a lane‑change scenario for an EIV moving on a two‑lane road of $7\,\textrm{m}\times3,000\,\textrm{m}$. Besides this EIV, $35$ HVs are uniformly distributed across this road according to the simulator settings. During the simulation, these HVs follow the built-in behavior model. We use the fine‑tuned PaliGemma model\footnote{\href{https://github.com/hanker-zhu/DriveVLM-project}{https://github.com/hanker-zhu/DriveVLM-project}.}, to provide the action recommendations. The PaliGemma model consists of a Transformer decoder and a Vision Transformer-based image encoder, with a total of 3 billion parameters. The text decoder is initialized with weights from the Gemma 2B model, and the image encoder uses a pre-trained SigLIP So400m Patch14 architecture. We compare our framework with two baselines: \rmnum{1}) Integration of LMMs with the state-of-the-art DRL algorithms, including \textbf{DDQN} and \textbf{DQN}; \rmnum{2}) EI driving scheme without the assistance of LMMs (\textbf{W/O LMM}), where the semantic pipeline is not included, and the network structure of DRL approach remains identical as our framework.

\begin{figure*}[t]
	\centering
	\subfigcapskip=-6.1pt
	\hspace{-5mm}
	\subfigure[]{
		\begin{centering}
			\includegraphics[scale=0.45]{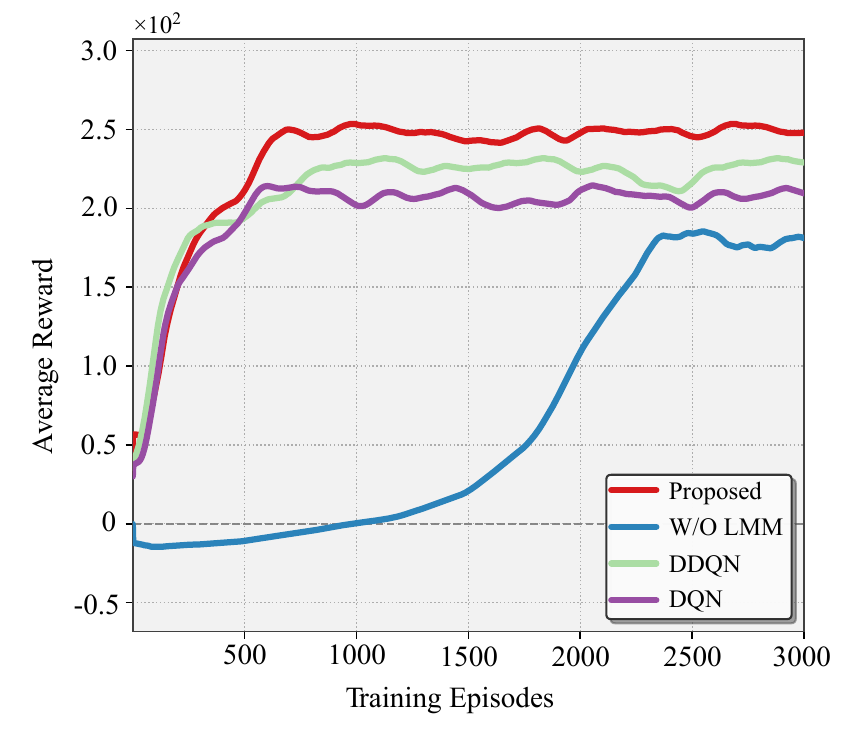}
		\end{centering}\hspace{0mm}
		\label{Fig-reward}}	
	\subfigure[]{
		\begin{centering}
			\includegraphics[scale=0.43]{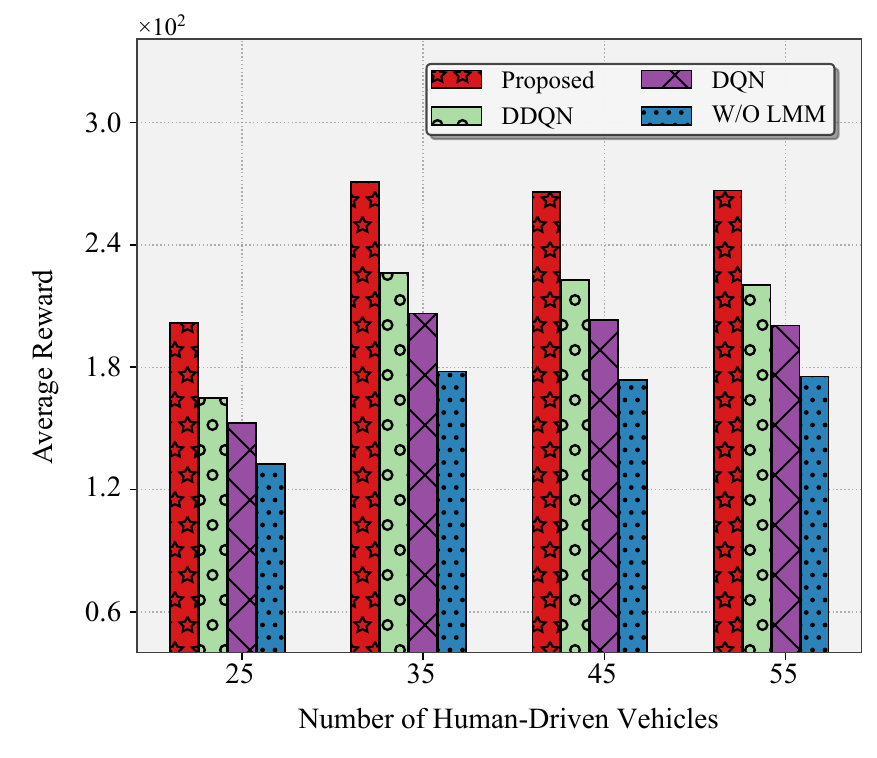}
		\end{centering}
		\label{Fig-reward-HVs}}
	\caption{Performance evaluation of the proposed dual-driven hybrid decision framework, compared with the baseline schemes including the state-of-the-art DRL algorithms and the particular case of W/O LMM: (a) Convergence performance of the adopted D3QN algorithm integrated with LMMs in the proposed framework; (b) Average reward versus the number of HVs. During each training episode, the cumulative reward is obtained by summing the rewards over all interaction steps. The average reward is then calculated by averaging the episodic cumulative rewards over all training episodes. Across multiple independent runs, the average reward varies by approximately $\pm$6.}
	\label{Fig1-simulation-results}
\end{figure*}

\textbf{Dataset.}
Throughout the simulations, we employ the open-source nuScenes dataset\footnote{\href{https://www.nuscenes.org/nuscenes}{https://www.nuscenes.org/nuscenes}.}, which integrates the structured data from three modalities. These modalities include multi-view scene videos (local), bird's-eye-view (BEV) map images (global), and multi-round Question Answering (QA) annotations. The nuScenes dataset consists of images from 1,000 driving scenes, captured across diverse urban and highway environments. Besides, it includes 18,000 reasoning-based Question Answering pairs, and more than 50,000 reasoning steps. This ensures the model learns to generate structured rationales before arriving at a final answer. The nuScenes dataset is used to provide multimodal preliminary semantic knowledge for the LMM-based semantic pipeline, including scene understanding and high-level action recommendation.

\textbf{Training and Hyperparameters.}
We adopt LoRA-based fine-tuning, and utilize AdamW optimizer to reduce memory requirements and achieve rapid adaptation to driving semantics and actions. The fine-tuned model is integrated with the D3QN algorithm \cite{ZhangL2025}, for reducing erroneous interventions and outputting deterministic policy. The semantic pipeline is coupled with the reinforcement learning environment through the action recommendations generated by LMMs from the nuScenes data. The D3QN algorithm uses a four-layer fully connected neural network with two hidden layers, each containing 256 neurons and ReLU activation. The evaluation network updates its parameters via Adam optimizer. We set the reward-associated parameters as ${\delta_1} = -15$, ${\delta _2} = 10$, and ${\delta _3} = 2$, respectively. The simulation is configured with 3,000 training episodes, local batch size of 32, and learning rate of 0.001. 

\textbf{Results Evaluation.}
Fig. \ref{Fig-reward} presents the convergence performance of the adopted D3QN algorithm integrated with LMMs in our framework against the baselines. In DRL, rewards constitute the sole feedback signal during training and provide a direct measure of the policy quality.  As observed, our framework exhibits superior convergence performance and ultimately achieves a higher average reward than all baselines. Notably, our framework outperforms both the DDQN and DQN algorithms by achieving an average reward gain of at least 19.47\% and 31.07\%, respectively. This superiority stems from the Dueling structure of D3QN, which decouples the state values and action advantages. In contrast, DDQN only addresses Q-value overestimation, while DQN's single network structure is more prone to local optima, leading to limited exploration efficiency. We can also see that the W/O LMM scheme shows inferior convergence performance. This is because W/O LMM relies entirely on the algorithm for exploration. In the proposed framework, LMMs interpret the multimodal driving context and generate a high-level action recommendation, which serves as semantic guidance for policy decision. Such semantic guidance reduces uninformed exploration and guides the policy toward more appropriate actions. Consequently, our framework converges faster and achieves a higher average reward than the W/O LMM scheme.

Fig. \ref{Fig-reward-HVs} illustrates the impact of number of HVs on the average rewards for our framework and the baselines. As shown, our framework consistently outperforms all baselines, demonstrating superior dynamic adaptability. Particularly, all schemes exhibit a trend of initial increase followed by a decrease. As the number of HVs increases, the average reward rises gradually, peaking at 35 vehicles, before declining. This is because the EIV gains more lane-change opportunities due to the moderate increase in HVs, facilitating higher reward. However, an increasing number of HVs reduces driving maneuvering space, compelling the EIV to adopt conservative driving behavior to avoid collision, thereby limiting reward growth.

\section{Open Challenges and Future Directions}

\subsection{Virtual and Reality Collaborative Training}
The proposed framework has thus far been evaluated only in a simulated driving environment, while its adaptability to real-world traffic conditions remains to be investigated. Although the simulation enables efficient policy learning through extensive environment interactions, the discrepancy between simulated and real-world traffic may limit the generalization of the learned policy. Future research should exploit high-fidelity simulation environments and world models for large-scale pretraining and policy optimization, while incorporating real-world driving data for fine-tuning and adaptation, thereby improving policy generalization to complex scenarios.

\subsection{Real-Time and Efficient Model Deployment}
Integrating LMM-based semantic reasoning inevitably introduces additional inference latency and computational overhead compared with DRL policies. Although the policy pipeline can provide real-time executable actions, the computational efficiency of the semantic pipeline remains important for practical deployment. Future research should investigate lightweight multimodal models, model compression, and asynchronous semantic policy execution to reduce computational cost while preserving high-level reasoning capability.

\subsection{Intelligent Interaction with Human-Driven Vehicles}
We consider mixed traffic but the interactions among vehicles are still represented in a simplified manner. In realistic traffic environments, HVs exhibit heterogeneous and uncertain behaviors that can significantly affect lane-change decisions. Future research should incorporate more driver intention inference, behavior prediction, and interactive policy learning into the proposed framework. EIVs could better anticipate the responses of surrounding vehicles and achieve more cooperative and adaptive decision in mixed traffic.

\section{Conclusion}
In this article, we have proposed a semantics and policy dual-driven hybrid decision framework to achieve continuous learning and joint decision for enhancing EI driving. 
Our framework has merged LMMs, for semantic understanding and cognitive representation, and DRL, for real-time policy optimization.
We also have provided the emerging opportunities this framework empowers, from the potential benefits to the transformative use cases.
Furthermore, a case study, which aims for lane-change planning in an EI vehicle and conventional HV mixed scenario, has demonstrated the superiority of our framework in policy quality and dynamic adaptability.
Concluding the article, some open challenges and prospective research directions pertaining to EI driving have been outlined for future exploration.


\begin{thebibliography}{1}
	
\bibitem{ZhangL2025}
L. Zhang, T. Song, L. Li, L. Chen, D. Niyato, and Z. Han, ``Multimodal semantic communications empowered lane-change planning for autonomous driving,'' \textit{IEEE Trans. Veh. Technol.}, vol. 75, no. 2, pp. 3330--3335, Feb. 2026.

\bibitem{ClaussmannL2019}
L. Claussmann, M. Revilloud, D. Gruyer, and S. Glaser, ``A review of motion planning for highway autonomous driving,'' \emph{IEEE Trans. Intell. Transp. Syst.}, vol. 21, no. 5, pp. 1826--1848, May 2020.

\bibitem{ChenL2024}
L. Chen, O. Sinavski, J. H\"unermann, A. Karnsund, A. J. Willmott, D. Birch, D. Maund, and J. Shotton, ``Driving with LLMs: Fusing object-level vector modality for explainable autonomous driving,'' in \textit{Proc. IEEE Int. Conf. Robot. Automat.}, Yokohama, Japan, May 2024, pp. 14093--14100.

\bibitem{HuangS2025}
S. Huang, F. Shi, C. Sun, J. Zhong, M. Ning, Y. Yang, Y. Lu, H. Wang, and A. Khajepour, ``DriveSOTIF: Advancing SOTIF through multimodal large language models,'' \textit{IEEE Trans. Veh. Technol.}, vol. 75, no. 3, pp. 3642--3655, Mar. 2026.
	
\bibitem{LiaoH2025}
H. Liao, B. Rao, H. Sun, C. Wang, Q. Chang, S. E. Li, C. Xu, and Z. Li, ``Chain-of-thought guided multimodal large language models for scene-aware accident anticipation in autonomous driving,'' \emph{IEEE Trans. Intell. Transp. Syst.}, vol. 26, no. 1, pp. 19371--19380, Nov. 2025.

\bibitem{XuZ2024}
Z. Xu, Y. Zhang, E. Xie, Z. Zhao, Y. Guo, K.-Y. K. Wong, Z. Li, and H. Zhao, ``DriveGPT4: Interpretable end-to-end autonomous driving via large language model,'' \textit{IEEE Robot. Autom. Lett.}, vol. 9, no. 10, pp. 8186--8193, Oct. 2024.

\bibitem{HuC2025}
C. Hu and X. Li, ``Human-centric context and self-uncertainty-driven multi-modal large language model for training-free vision-based driver state recognition,'' \emph{IEEE Trans. Intell. Transp. Syst.}, early access, Apr. 2025, doi: 10.1109/TITS.2025.3558847.

\bibitem{DuanJ2022}
J. Duan, S. Yu, H. L. Tan, H. Zhu, and C. Tan, ``A survey of embodied AI: From simulators to research tasks,'' \textit{IEEE Trans. Emerg. Top. Comput. Intell.}, vol. 6, no. 2, pp. 230--244, Apr. 2022.

\bibitem{ZhangR2025}
R. Zhang, C. Zhao, H. Du, D. Niyato,  J. Wang, S. Sawadsitang, X. Shen, and D. I. Kim, ``Embodied AI-enhanced vehicular networks: An integrated vision language models and reinforcement learning method,'' \emph{IEEE Trans. Mobile Comput.}, vol. 24, no. 11, pp. 11494--11510, Nov. 2025.

\bibitem{ChenM2025}
M. Chen, C. Wang, X. He, F. Zhu, L. Wang, and A. V. Vasilakos, ``Embodied artificial intelligence-enabled internet of vehicles: Challenges and solutions,'' \textit{IEEE Veh. Technol. Mag.}, vol. 20, no. 2, pp. 63--70, Jun. 2025.

\bibitem{ZhouM2024}
M. Zhou, H. Dong, H. Song, N. Zheng, W.-H. Chen, and H. Wang, ``Embodied intelligence-based perception, decision-making, and control for autonomous operations of rail transportation,'' \textit{IEEE Trans. Intell. Veh.}, vol. 10, no. 12, pp. 5061--5065, Dec. 2025.

\bibitem{LiL2025}
L. Li, Y. Li, X. Zhang, Y. He, J. Yang, B. Tian, Y. Ai, L. Li, A. N\"uchter, and Z. Xuanyuan, ``Embodied intelligence in mining: leveraging multi-modal large language models for autonomous driving in mines,'' \textit{IEEE Trans. Intell. Veh.}, vol. 9, no. 5, pp. 4831--4834, May 2024.

\bibitem{ChengL2025}
L. Cheng, H. Zhang, B. Di, D. Niyato, and L. Song, ``Large language models empower multimodal integrated sensing and communication,'' \textit{IEEE Commun. Mag.}, vol. 63, no. 5, pp. 190--197, May 2025.

\bibitem{ZhangR2023}
R. Zhang, K. Xiong, Y. Lu, P. Fan, D. W. K. Ng, and K. B. Letaief, ``Energy efficiency maximization in RIS-assisted SWIPT networks with RSMA: A PPO-based approach,'' \textit{IEEE J. Sel. Areas Commun.}, vol. 41, no. 5, pp. 1413--1430, 2023.

\bibitem{ZhangL2025B}
L. Zhang, Z. Wu, H. Xu, D. Niyato, C. S. Hong, and Z. Han, ``Digital twin-driven federated learning for converged computing and networking at the edge,'' \textit{IEEE Netw.}, vol. 39, no. 2, pp. 20--28, Mar. 2025.



\end{thebibliography}
\end{document}